\documentclass[a4paper, 10pt, journal]{orbieeeconfpost}      
\conference{IEEE-RAS International Conference on Humanoid Robots (Humanoids 2016)}

\IEEEoverridecommandlockouts                              

\usepackage{graphicx}
\usepackage{tabularx}
\usepackage{booktabs}
\usepackage{graphics} 
\usepackage{amsmath} 
\usepackage{amssymb}  
\usepackage{caption}
\usepackage{subcaption}
\usepackage{siunitx}
\title{\LARGE \bf
Walking of the iCub humanoid robot in different scenarios: implementation and performance analysis
}

\author{Yue Hu$^{1}$, Jorhabib Eljaik $^{2}$, Kevin Stein$^{1}$, Francesco Nori$^{2}$, Katja Mombaur$^{1}$
\thanks{The research leading to these
results has received funding from the European Union Seventh Framework
Programme (FP7/2007 - 2013) under grant agreement n. 611909 (KoroiBot), www.koroibot.eu}
\thanks{$^{1}$Interdisciplinary Center for Scientific Computing, Heidelberg University, 69115 Heidelberg, Germany}%
\thanks{$^{2}$Department of Robotics, Brain and Cognitive Sciences, Fondazione Istituto Italiano di Tecnologia (IIT), 16163 Genoa, Italy}
\thanks{
        {\tt\small yue.hu@iwr.uni-heidelberg.de}}%
\thanks{
        {\tt\small jorhabib.eljaik@iit.it}}%
\thanks{
        {\tt\small kevin.stein@iwr.uni-heidelberg.de}}%
\thanks{
        {\tt\small francesco.nori@iit.it}}%
\thanks{
        {\tt\small katja.mombaur@iwr.uni-heidelberg.de}}%
}

\begin{document}

\maketitle

\begin{abstract}
The humanoid robot iCub is a research platform of the Fondazione Istituto Italiano di Tecnologia (IIT), spread among different institutes around the world.
In the most recent version of iCub, the robot is equipped with stronger legs and bigger feet, allowing it to perform balancing and walking motions that were not possible with the first generations.
Despite the new legs hardware, walking has been rarely performed on the iCub robot. In this work the objective is to implement walking motions on the robot, from which we want to analyze its walking capabilities.
We developed software modules based on extensions of classic techniques such as the ZMP based pattern generator and position control to identify which are the characteristics as well as limitations of the robot against different walking tasks in order to give the users a reference of the performance of the robot. Most of the experiments have been performed with HeiCub, a reduced version of iCub without arms and head.
\end{abstract}

\section{INTRODUCTION}
Being humanoid robots still at a research level, there are only few platforms that are widely spread and studied across research institutes.
The Fondazione Istituto Italiano di Tecnologia (IIT) built the robot iCub \cite{metta2010icub} and it is now available for research purposes in several institutes all over the world in different versions.

The iCub is a medium size child-like humanoid robot with 53 degrees of freedom (DOF), of which 38 DOF in the upper body including 18 DOF of the hands, and 15 DOF in the torso and legs. It has a height of circa 1.1\si{\meter} and weight of 33\si{\kilogram}. It is a very complex system designed with the aim of performing cognitive science studies \cite{metta2010icub}.
Among the current abilities of iCub there are objects recognition, voice commands recognition, grasping, tactile sensing, and many others that are growing thanks to the several projects in which it is involved and the open source quality of the project.

In the first versions, iCub had very small feet as it was not designed for walking. In recent versions of iCub, bigger feet were introduced, allowing the robot to perform whole body balancing tasks in double and single support \cite{iCubWBC}.
In recent hardware upgrades the mechanical design of the legs were derived from the CoMAN humanoid robot of IIT \cite{colasanto2012compact}. The new legs have Series Elastic Actuators (SEA) with removable springs in the knee and ankle pitch joints \cite{parmiggiani2012mechatronic}.

In the framework of the European Project KoroiBot \cite{koroiweb}, a customized version of the iCub was delivered to Heidelberg University (hereafter HeiCub) in order to perform walking experiments. The HeiCub robot is an iCub without head and arms and with only 15 DOF, namely 3 in the torso and 6 in each leg. It has the standard iCub hardware \cite{metta2010icub}, including two fixed frontal cameras mounted in the torso and the SEAs with removable springs. In the scope of this work the springs are removed to use rigid actuators only.

\begin{figure}
\begin{center}
 \includegraphics[scale = 0.12]{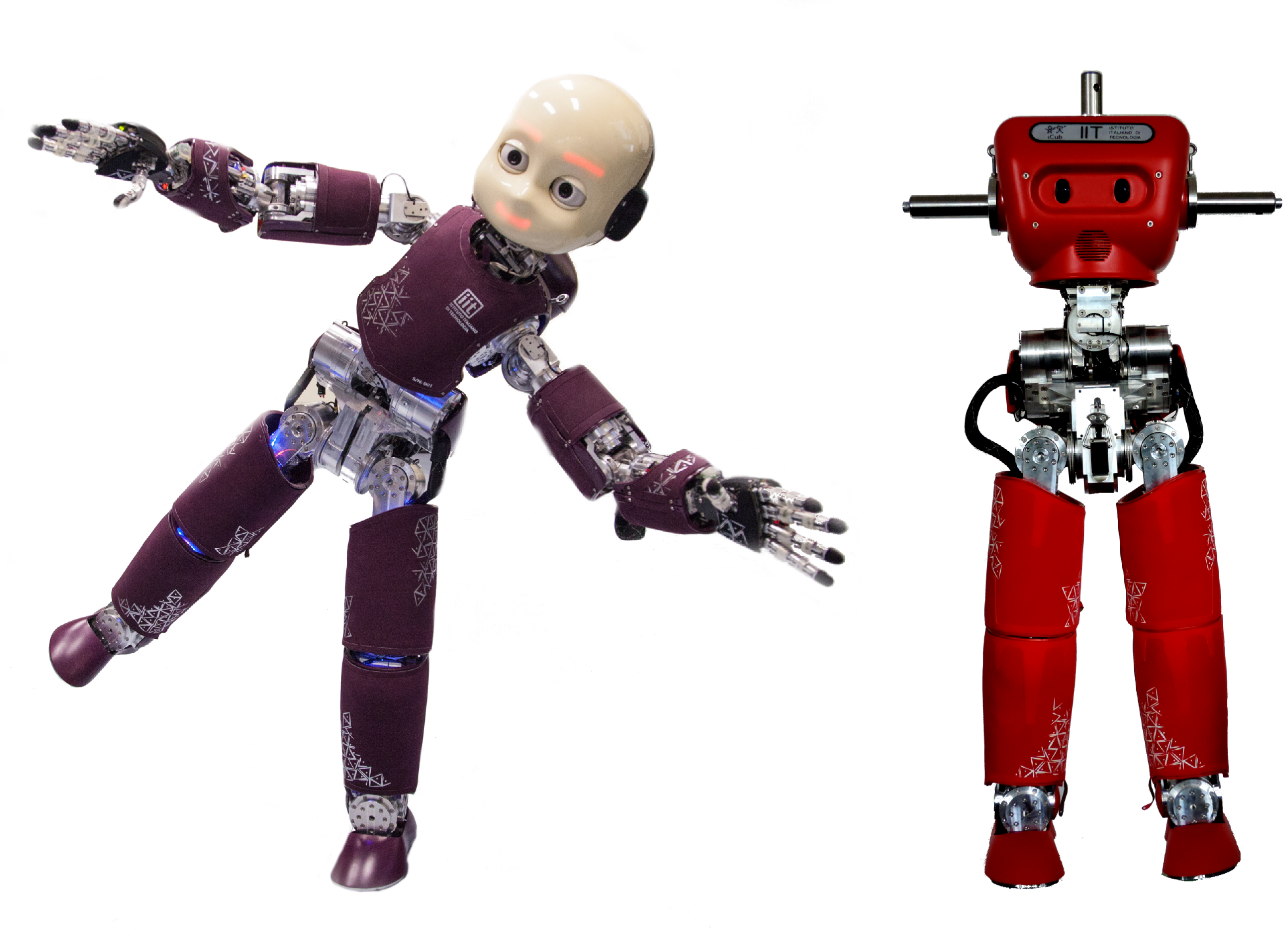}
 \caption{\small{iCub and HeiCub robots, designed and built by iCub Facility department, IIT}}
\end{center}
\end{figure}

The objective of the KoroiBot Project is the improvement of existing humanoid robots walking abilities. 
In the project we first analyze the current walking skills of the available platforms to have a comparison between the actual performances and those that will come with the novel methods provided by the project.

On some of the robots of the project, however, walking has never or rarely been implemented, either in simulation or on the actual platform.
This is the case of iCub, on which there have been only few experiments on walking on level ground, despite being such a diffused robot in research facilities as well as projects.

With this work we want to bring walking motions on the iCub robot in different scenarios for the first time.
To this end we developed software modules to implement walking motions on the robot in order to identify suitable environmental constraints within its mechanical limits and perform quantitative measurements, e.g. tracking precision and current consumption.
The objective is to give the iCub community as well as the humanoid research community a point of reference for the possibilities and limitations of the iCub in walking using a simple software module to generate walking motions.

In order to achieve our goal, we implemented an offline ZMP based pattern generator, which is an extension of the work in \cite{kajita2003biped}, that generates feet and center of mass (CoM) trajectories, which are then given as inputs to a module that performs inverse kinematics to retrieve joint trajectories. The obtained trajectories are executed on the robot using position control.

The approach has been used to carry out walking in different scenarios, including level ground, slopes and stairs. In every scenario different patterns have been tested to reach the limits of the robots within the proposed methods, where we also analyze the reason of the unsuccessful trials. The experiments were performed mainly on the HeiCub robot, but in the flat ground case also on complete iCub(s) at IIT.

\section{SOFTWARE MODULES}
The software module consists of three parts: the pattern generator, the inverse kinematics and the interface to the robot.
The pattern generator takes as input user defined foothold positions for right and left feet and a predefined center of mass height variation pattern that is adapted to different environments, from which it generates the feet trajectories and the 3D center of mass trajectory. These trajectories are then used by inverse kinematics as desired end effector positions to compute the joint trajectories. The center of mass is assumed as a point attached to the chest of the robot. In order to match the real center of mass, the coordinates of this point are updated with the real center of mass at every time instant, instead of assuming it being a fixed point attached to a link (usually the pelvis) as many applications do.

The reason for which we introduced the CoM height variation is to obtain motions that are closer to human-like walking, where it is known that the CoM height has a wave shape variation. This variation allows the robot to walk with stretched knee and perform bigger steps despite restricted joint limits as we will discuss later in section \ref{sec:res}.

In the following sections we review the theory behind the first two parts, explaining also the extensions introduced to fulfill our goals. Then the full workflow is described.

\subsection{Pattern generator}
\label{subsec:pg}
One of the most common methods used in humanoid robotics to generate walking motions is the inverted pendulum or table cart combined with the ZMP principle \cite{vukobratovic2004zmp}. Here we make an extension on the one proposed by Kajita et al. in \cite{kajita2003biped}, which allows arbitrary foot placement and assumes fixed center of mass height over the whole walking. In our extension we allow the center of mass height to vary according to a predefined pattern.

The pattern generator is based on 3D linear inverted pedulum and ZMP as in \cite{kajita2003biped} and \cite{Kajita2002}, where the ZMP equations including CoM height variation are described as:
\begin{equation}
 \mathbf p = \mathbf x - \frac{z\mathbf{\ddot x}}{\ddot z + g}
\label{eq:zmp}
\end{equation}
Where $\mathbf p = [p_x, p_y]$ represents the ZMP coordinates, $\mathbf x = [x,y]^T$ the CoM coordinates, $z$ the height variation of the CoM and $g$ the gravity.

In \cite{kajita2003biped} a constrained plane $z = k_xx+k_yy+z_c$ is assumed. By imposing $k_x = k_y = 0$, the equation is simplified into:
\begin{equation}
 \mathbf p = \mathbf x - \frac{z_c}{g}\mathbf {\ddot x}
\end{equation}
However in this work we use the formulation as in equation \eqref{eq:zmp}, as we want to set a predefined variable CoM trajectory in the $z$ direction instead of keeping the CoM height constant.

The CoM on the walking plane ($x$ and $y$ directions) is computed from equation \eqref{eq:zmp} as described in \cite{kajitabook}, with modifications introduced to take into account the $z$ variations.
The accelarations $\ddot x$ is approximated as:
\begin{equation}
 \ddot x_i = \frac{x_{i-1}-2x_i+x_{i+1}}{\Delta t^2}
\end{equation}
where $i = 1...N$ is the number of the sample with $N$ being the total number of samples. The same approximation is done for $\ddot y$.

The ZMP equation as in Eq. \eqref{eq:zmp} can be discretized as follows to have the formulation $\mathbf p = \mathbf A\mathbf x$ (the same is applied for the $y$ coordinate):
\begin{equation}
 \begin{bmatrix}
  p_1\\p_2\\ \vdots\\p_{N-1}\\p_N
 \end{bmatrix} =
 \begin{bmatrix}
  a_1+b_1 & c_1 & 0 &
  \\
  a_2 & b_2 & c_ 2 &
  \\
  & \ddots & \ddots & \ddots  &
  \\
   & a_{N-1} & b_{N-1} & c_{N-1}
  \\
   & 0 & a_N & b_N+c_N
 \end{bmatrix}
 \begin{bmatrix}
  x_1 \\ x_2 \\ \vdots \\ x_{N-1} \\ x_N
 \end{bmatrix}
\end{equation}
Where the terms of the matrix are as follows:
\begin{equation}
\begin{array}{l}
 a_i = -\dfrac{z}{(\ddot z + g)\Delta t^2}
 \\
 b_i = \dfrac{2z}{(\ddot z + g)\Delta t^2} + 1
 \\
 c_i = a_i
\end{array}
\end{equation}
The system is then solved for the CoM coordinates $\mathbf x$.
The matrix $\mathbf A$ has dimension $N\times N$. This means that the higher is the number of samples, the bigger is the matrix $\mathbf A$. However this computation is performed offline and does not represent a concern at the moment.

\subsection{Inverse kinematics (IK)}
In order to perform walking in different environments, we need inverse kinematics that gives feasible joint angles for given multiple end effectors positions and orientations.
Therefore we extended and modified the existing algorithm implemented in the Rigid Body Dynamics Library (RBDL) \cite{Felis2016}, which did not allow orientation constraints.
We adopt the residuals definition as of Sugihara in \cite{Sugihara:LMM}:
\begin{equation}
 \mathbf e_i(\mathbf q) = \begin{cases}
           ^d\mathbf p_i-\mathbf p_i(\mathbf q)\\\mathbf R_i^T(\mathbf q)\cdot a(^d\mathbf R_i\mathbf R_i^T(\mathbf q))
          \end{cases}
\end{equation}
Where $^d\mathbf p_i$ and $\mathbf p_i$ are the desired and current positions of the end effectors $i$ and $a(\cdot)$ is a function that computes angular velocities from the current rotation matrix \footnote{It is also worth noting that the expression of the orientation error depends on the convention adopted for the rotation matrices, the original formulation from Sugihara \cite{Sugihara:LMM} has been changed to a different convention adopted in RBDL.} $\mathbf R_i$ and the desired orientation expressed as rotation matrix $^d\mathbf R_i$.

The openly available software for solving non-linear optimization problems IPOPT \cite{ipopt} has been used to solve the inverse kinematics problem, which is formulated as follows:
\begin{equation}
\begin{array}{rrclcl}
\displaystyle \min_q & \multicolumn{3}{l}{\lVert\mathbf e(\mathbf q)\rVert^2_2}
\end{array}
\end{equation}
subject to:
\begin{equation}
\begin{array}{rclcl}
\mathbf q & \geq & \mathbf q_{min}\\
\mathbf q & \leq & \mathbf q_{max}
\end{array}
\label{Eq:OptimContConstraints}
\end{equation}
Where $\mathbf e$ is the residuals of stacked position and orientation errors of all end effectors and $\mathbf q_{min}$ and $\mathbf q_{max}$ are the vectors representing the minimum and maximum limits of every joint.

The following quantities are defined in order to solve the optimization problem using the Ipopt solver:
\begin{itemize}
 \item Gradient of the objective: $\nabla \mathbf e = \mathbf J^T\mathbf e$.  \newline Where $\mathbf J$ is the the Jacobian of all the end effectors stacked $\mathbf J = \{\mathbf J_0,...,\mathbf J_N\}$. The single $\mathbf J_i$ describes $^0\mathbf{\dot X}_i = \mathbf J_i\mathbf q$, being $^0\mathbf{\dot X}_i \in \Re^6$ the linear and angular velocities of the end effector $i$ in the world reference frame.
 \item Hessian of the Lagrangian function, for which we use a Gauss-Newton approximation: $\mathbf H \approx \mathbf J^T\mathbf J$, where $\mathbf J$ is the same Jacobian as in the gradient.
\end{itemize}
In common optimization problems the Jacobian of the constraints would also be required, but since in our case the constraints consist only in the box constraints represented by the joint limits given that the task of matching the end effector positions are formulated in the objective function, we do not need to formulate the constraints Jacobian.

In the implementation we allow also to use individual position or orientation constraints, as this is useful in some cases, e.g. in circle walking, where by constraining the feet and chest only the pelvis would not rotate with the body, in this case an additional orientation constraint on the pelvis can be imposed.

\subsection{Workflow}
\label{sec:workflow}
The pattern generator is implemented in MATLAB (The MathWorks, Inc.), it takes as input a file containing several parameters for the different walking environments.
The script automatically generates the feet pattern (foothold positions for right and left feet) for a given set of parameters as in Table \ref{tab:pgparams}.

\begin{table}
  \begin{center}
  \caption{Pattern generator parameters}
  \begin{tabular}{|c|c|}
    \hline
    \textbf{Name} & \textbf{Description}\\
    \hline
    ts & sampling time \\
    \hline
    z\_c & CoM height \\
    \hline
    z\_c\_offset & height variation of CoM \\
    \hline
    n\_strides & number of strides\\
    \hline
    T\_stride & time to perform 1 stride \\
    \hline
    T\_switch & double support time \\
    \hline
    step\_width & distance between two feet \\
    \hline
    step\_length & distance traversed during 1 step \\
    \hline
    theta & inclination during slope walking \\
    \hline
    stair\_length & length of the stair \\
    \hline
    stair\_height & height of the stair \\
    \hline
    right\_step\_first & by default the first to step is the left foot \\
    \hline
    type & specifies the walking environment\\
    \hline
  \end{tabular}
  \label{tab:pgparams}
  \end{center}
\end{table}

From the feet pattern the desired ZMP trajectory ZMP$_d$ is computed as well as the desired trajectory of the CoM$_z$.
The ZMP trajectory is generated such that a smooth transition of the ZMP from one foot to the other is performed. This corresponds to having a double support phase of the duration T\_switch, which is included in the stride time. This means that the time to perform a step, in the sense of leg swing from one foot hold to the next one, corresponds to (T\_stride$-$T\_switch)/2.

The desired CoM height variation CoM$_z$ is computed from a set of heights given at desired times, then a spline interpolation is performed to obtain the CoM$_z$ trajectory. As explained previously, this is inspired from human walking and allows the robot to perform larger motions. In the case of level ground walking we compose the pattern by introducing a height offset corresponding to half time of the single support time, such that the robot stretches the stance leg during this phase and then bends it to go towards the double support phase. In the case of stairs the height offset is introduced in correspondence to when the foot has to step over the stair, such that the robot can clear the edge of the stair by stretching the stance leg and impose less force on the knee joint of the stance leg.

The ZMP$_d$ and CoM$_z$ are given as inputs to a function that implements the theory described in section \ref{subsec:pg}, obtaining the full CoM trajectory CoM$_d$.
Feet trajectories are generated using spline interpolation, by introducing a certain height (step\_height) to lift the feet off the floor during the swing phase.

The computed CoM and feet trajectories are given as inputs to the inverse kinematics module implemented in C++, where the kinematic model of the robot is loaded from a robot specific URDF (Unified Robot Description File).
We use as end effectors the center of the sole reference frames of iCub and the CoM attached to the chest of the robot, at the end of the torso chain, in order to use also the torso joints to compensate for CoM variations. The local CoM is updated to the real CoM at every sample in order to match the correct CoM and have higher stability.

Once the joint trajectories are computed from inverse kinematics, we use an iCub-dedicated C++ software module to execute the motion on the robot through Yarp \cite{yarp} interfaces. The module first brings the robot from its current pose to the initial commanded one using a minimum jerk position controller, then sends joint position commands at a desired thread rate, which should correspond to the sampling time given in the parameters table.
The same workflow can be applied both for the iCub in the simulation environment and the real robot. Furthermore the pattern generator and inverse kinematics module can be used with any two legged robot with an available UDRF file containing kinematic and dynamic properties.

\begin{figure}
 \begin{center}
  \includegraphics[scale = 0.5]{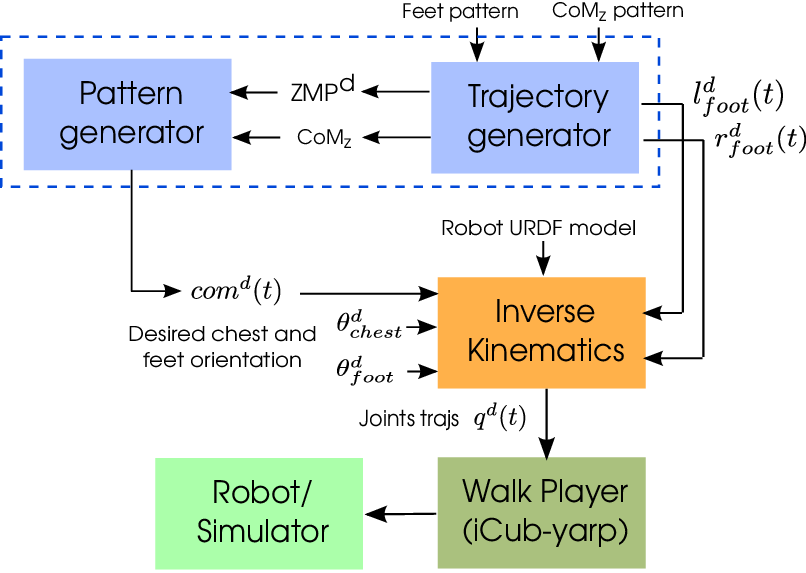}
  \caption{\small{Workflow from pattern generator to the robot}}
 \end{center}
\end{figure}

The following walking cases are supported:
\begin{itemize}
 \item Level ground straight walking
 \item Slope walking up/down
 \item Stairs climbing
\end{itemize}
It is also possible to generate ZMP and CoM trajectories for a given set of feet patterns that is not regular, e.g. step stones, as the algorithm is suitable for arbitrary foot placement. However this has not been included in this paper.

\section{RESULTS}
\label{sec:res}
We have applied the illustrated method to generate walking in the cases as per Table \ref{tab:env}. All the cases illustrated hereafter are referred to experiments on the HeiCub robot.
In all the cases we try to push the limits of the robot allowed by our method. Every case has been tested for 5 times to ensure repeatability of the motion. Specific cases will be illustrated hereafter.

It should be noted that walking down stairs has not been performed due to restricted joint limits of the ankle pitch joints. When walking down the stairs, the ankle needs a very wide angle even in humans, where it reaches up to -40$\sim$-45 \si{degree}. It needs to be higher in humanoid robots with flat feet such as the iCub, as there is no possibility of performing tip-toe rolling motions.

\begin{table}
 \begin{center}
 \caption {Walking environments and success rates over 5 trials for the HeiCub robot}
  \begin{tabular}{|c|c|c|c|}
  \hline
   Level ground & T\_stride $[s]$ & Step length $[m]$ & Success rate\\
  \hline
    & 8 & 0.10 & 100 \%\\
   \hline
   & 6 & 0.10 & 100 \%\\
   \hline
   & 5 & 0.10 & 100 \%\\
   \hline
   & 4 & 0.10 & 100 \%\\
   \hline
   & 3 & 0.10 & 0 \%\\
   \hline
   Slope up & T\_stride $[s]$ & Slope inclination $[\deg]$ & Success rate\\ 
   \hline
   & 8 & $ 4.5 $ & 100 \%\\
   \hline
    & 8 & $ 7 $ & 100 \%\\
    \hline
    & 6 & $ 7 $ & 100 \%\\
    \hline
    & 5 & $ 7 $ & 40 \%\\
    \hline
    Slope down & T\_stride $[s]$ & Slope inclination $[\deg]$ & Success rate\\ 
   \hline
   & 8 & $ 4.5 $ & 100 \%\\
   \hline
    & 8 & $ 7 $ & 100 \%\\
    \hline
    & 6 & $ 7 $ & 20 \%\\
    \hline
    & 5 & $ 7 $ & 0 \%\\
    \hline
   Stair up & T\_stride $[s]$ & Stair height $[m]$ & Success rate\\
   \hline
   & 10 & 0.01 & 100 \%\\
   \hline
    & 10 & 0.02 & 100 \%\\
   \hline
    & 8 & 0.02 & 100 \%\\
   \hline
  \end{tabular}
 \label{tab:env}
 \end{center}
\end{table}

\begin{figure}
 \begin{center}
  \includegraphics[scale = 0.7]{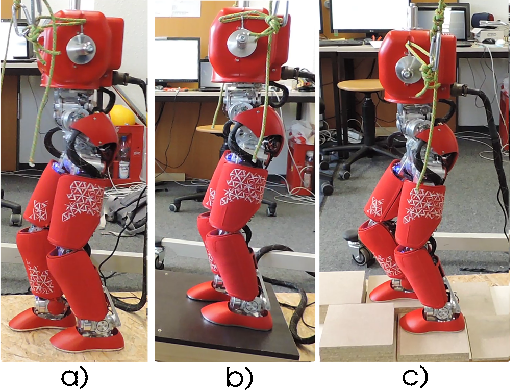}
  \caption{\small{Tested environments for HeiCub: level ground, slope (up and down), stairs.}}
 \label{Fig:scenario}
 \end{center}
\end{figure}

\subsection{Key Performance Indicators definition}
In the KoroiBot project, a series of performance indicators have been defined as a means to measure the capabilities of a robot. They are called the Key Performance Indicators (KPIs)\cite{kpi} and are divided in different categories such as human likeness, computational indicators and technical indicators.
We use here part of the technical KPIs as a means to summarize the performance of the iCub, which are defined as follows.
\begin{itemize}
 \item Cost of transport
 \begin{equation}
  \mathbf E_{CT} = \frac{\sum_{m=1}^{M} \int_{t_0}^{t_f} I_m(t)V_m(t)dt}{m_{robot}\cdot g \cdot d}
 \end{equation}
 Where $M$ is the total number of motors, in this case $M=15$, $I_m$ and $V_m$ are the current and voltage measurements of the motor $m$,  $m_{robot}$ is the mass of the robot, $g$ is the gravity and $d$ is the travelled distance.
 \item Froude number, used for level ground only
 \begin{equation}
  Fr = v_{max}/\sqrt{g\cdot h}, \ \text{for} \ h = {l_{leg}}
 \end{equation}
 \item Precision of task execution: we define them here as the set of tracking errors\footnote{Please note that quantities such as CoM and ZMP depend on the floating base estimation, which is now implemented in a module that performs the estimation using an odometry based method and the kinematic model of the robot, which is not highly accurate. For this reason the CoM and ZMP errors are not very precise and might be smaller than they result in this paper.} of the CoM, ZMP and joint angles, computed as RMSE (root mean squared error). Of these the most important is the ZMP error, as if the ZMP falls outside the support polygon the robot becomes unstable and might fall.
\end{itemize}
A table summarizing the KPIs of each case is reported in each section hereafter.

\subsection{The platform}
The illustrated experiments have been performed mainly on the HeiCub robot, but the workflow is perfectly applicable to all the existing iCub(s) with legs.
The difference between the robots lies mainly in the position of the CoM, which is computed from the dynamic parameters of the robot, described in a robot specific URDF (Unified Robot Description File).
From the kinematic point of view the HeiCub and all the other iCub(s) of the same generation are the same.
In particular, we have succesfully tested flat ground walk also with two full iCub(s) at the IIT facilities, one full size iCub with all the DOF and a full iCub with battery pack, which adds an additional weight of circa 3\si{\kilogram}.

The HeiCub robot, being without arms and head, weights 26.4\si{\kilogram}. The total height is 0.97\si{\meter}, the leg length (from the hip axis) is 0.51\si{\meter}, and the feet are 0.2\si{\meter} long and 0.1\si{\meter} wide, which are quite small compared to many other state of the art walking humanoid robots. The robot is equipped with a large variety of sensors that include joint and motor encoders, 6 axis force torque sensors, an inertial measurement unit (IMU) and many others, from which we gathered data for the performance analysis.

The standard iCub has restricted joint limits in the leg joints imposed via low level control, which do not correspond to the mechanical limits of the robot. We have extended these limits by leaving only a small margin for protection reasons. The limits are described as in Table \ref{tab:lim}.

Furthermore on the iCub a current limit of 5\si{\ampere} is imposed as protection for the motors. This limit is however too conservative for some walking tasks, such as large steps and stair walking. It has been increased on the HeiCub, however, this limit cannot be increased over a certain maximum, which is still not sufficient for more challenging tasks (e.g. higher stairs).

\begin{table}
 \begin{center}
 \caption{HeiCub joint limits}
  \begin{tabular}{|c|c|}
  \hline
  \textbf{Joint} & \textbf{Limits} [$\deg$] \\
  \hline
   l\_hip\_pitch, r\_hip\_pitch & [-33, 100] \\
   \hline
   l\_hip\_roll, r\_hip\_roll & [-19, 90] \\
   \hline
   l\_hip\_yaw, r\_hip\_yaw & [-75, 75] \\
   \hline
   l\_knee, r\_knee & [-100, 0] \\
   \hline
   l\_ankle\_pitch, r\_ankle\_pitch & [-36, 27]\\
   \hline
   l\_ankle\_roll, r\_ankle\_roll & [-24, 24]\\
   \hline
   torso\_pitch & [-20, 60]\\
   \hline
   torso\_roll & [-26, 26]\\
   \hline
   torso\_yaw & [-50, 50]\\
   \hline
  \end{tabular}
 \label{tab:lim}
 \end{center}
\end{table}

\subsection{Level ground}
\begin{figure}[!tbp]
    \centering
    \begin{subfigure}[b]{0.5\textwidth}
      \includegraphics[scale = 0.2]{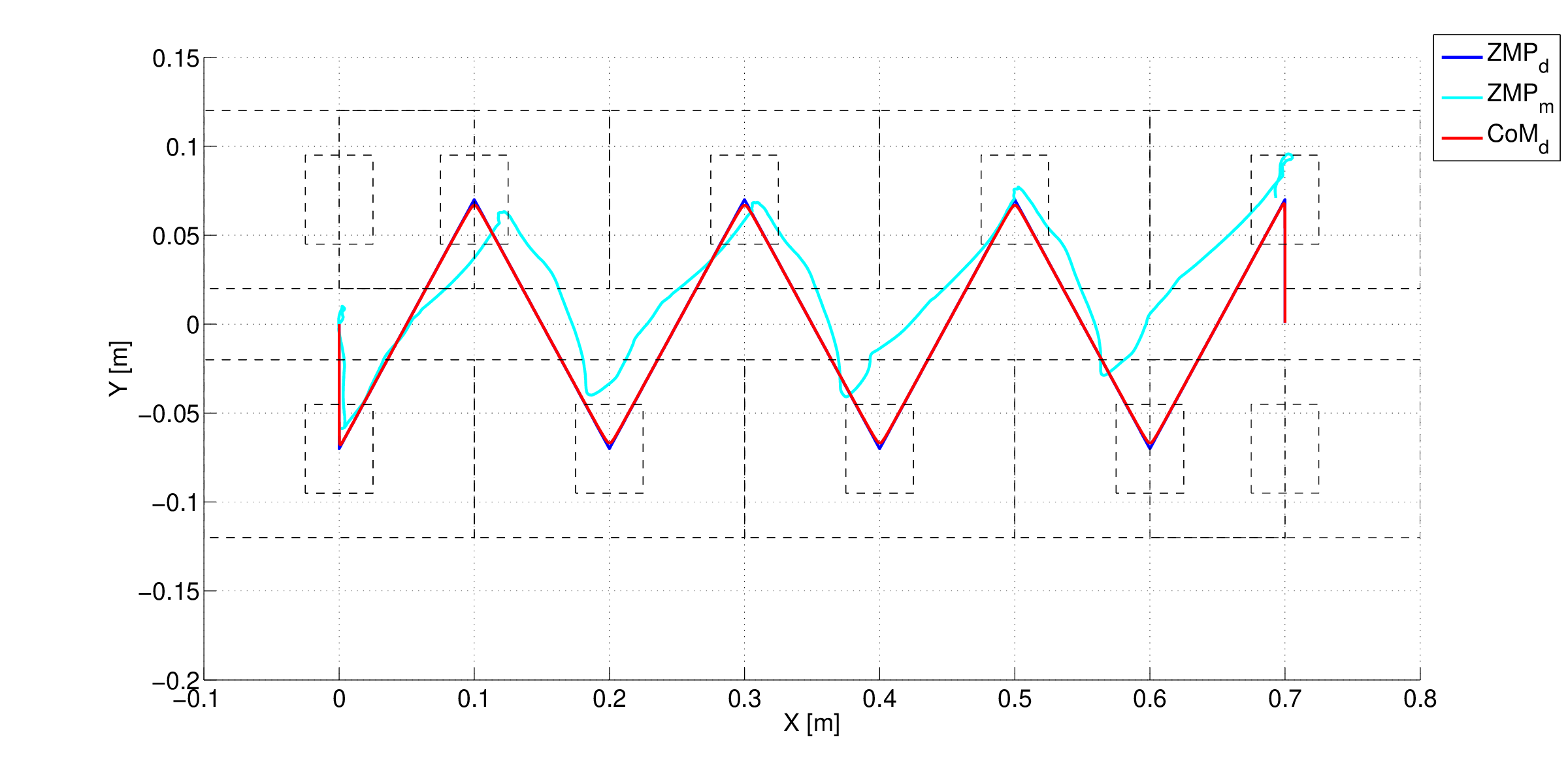}
      \caption{Feet pattern with ZMP and CoM}
      \label{Fig:level_pattern}
    \end{subfigure}

    \begin{subfigure}[b]{0.5\textwidth}
      \includegraphics[scale = 0.15]{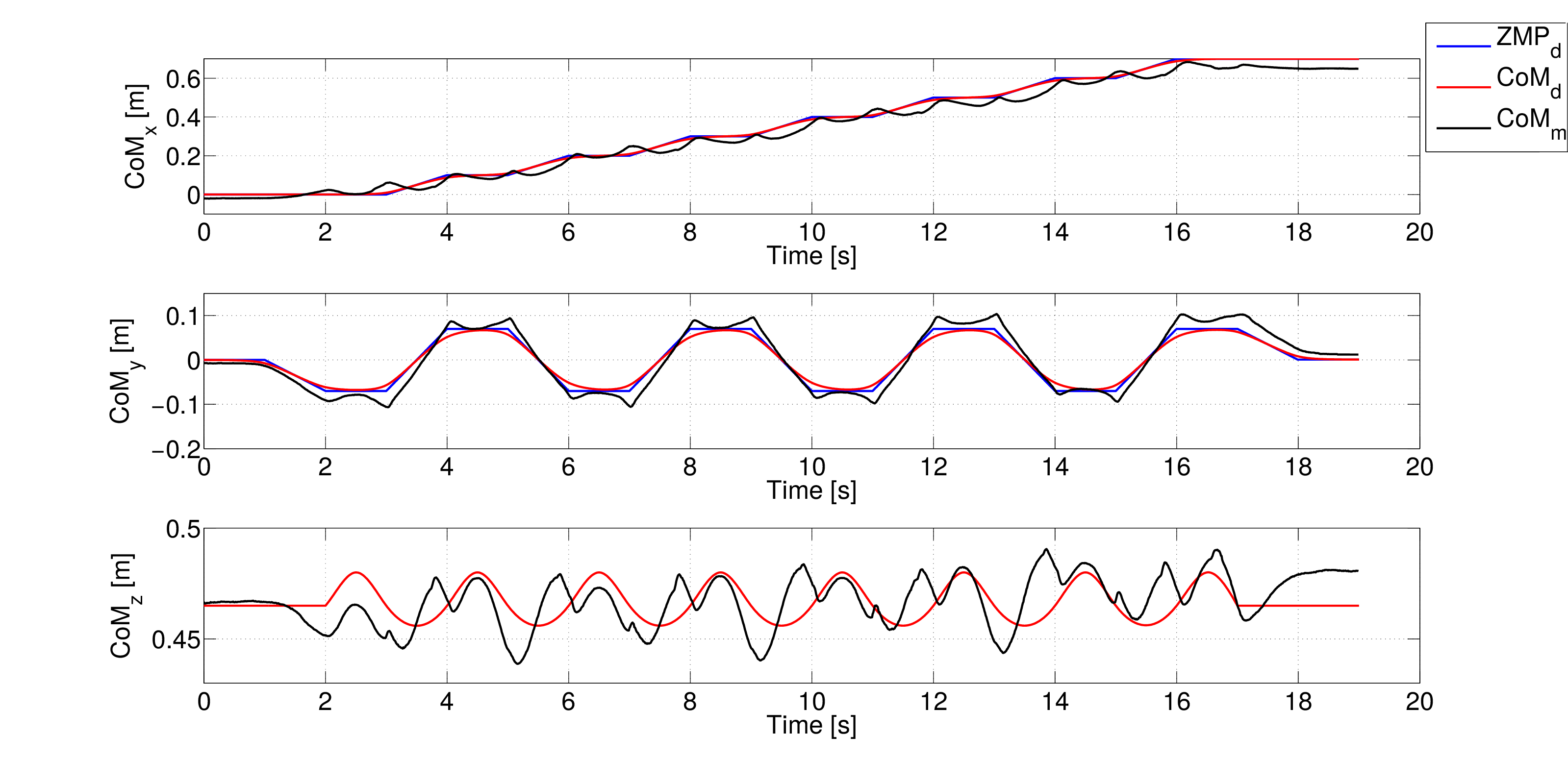}
      \caption{ZMP and CoM trajectory}
      \label{Fig:level_com}
    \end{subfigure}

    \begin{subfigure}[b]{0.5\textwidth}
      \includegraphics[scale = 0.2]{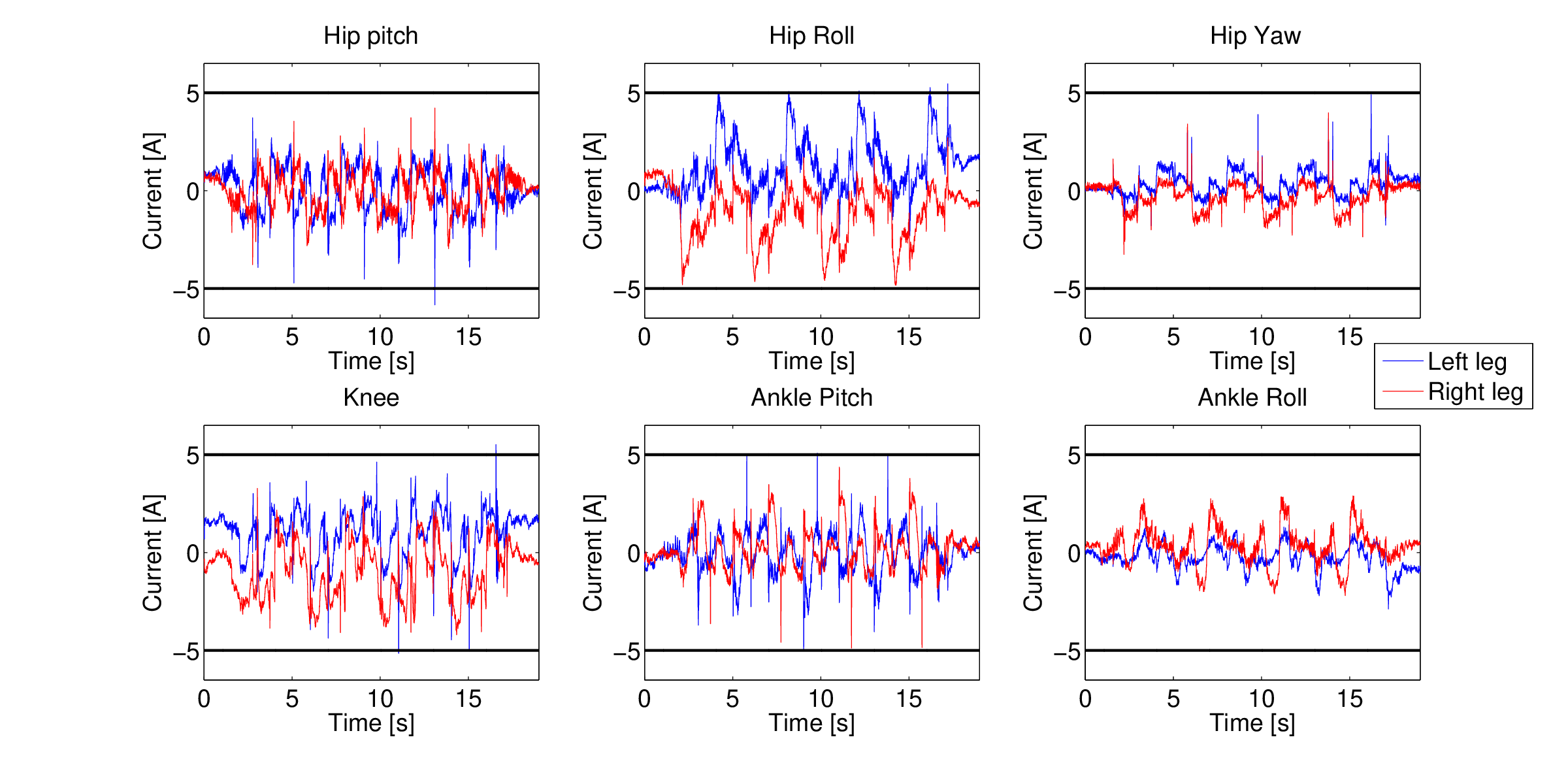}
      \caption{\small{Currents of leg motors, black lines are the default current limits of +/-5 [A].}}
      \label{Fig:level_current}
    \end{subfigure}
    \caption{\small{Walking on flat ground with T\_stride 4 [s] and step\_length 0.1 [m]. Center of mass has a variation of 0.015 [m].}}
    \label{Fig:level}
\end{figure}
Level ground walk has been performed with several different walking velocities. The step length was kept constant at 0.1\si{\meter} while the time has been reduced, as shown in Table \ref{tab:env}.

In Fig. \ref{Fig:level} we can see the plot of the feet pattern with the ZMP and CoM trajectories in the $[x,y]$ plane in the fastest case we could achieve, i.e. 4\si{\second} per stride.
In this case we used 1\si{\second} as double support time, which means that the swing time of a leg in single support phase is of 1.5\si{\second}.
Also, the step length is referred as the distance between the two feet when they are both on the ground, this means that the actual distance travelled by the foot in swing phase is twice the step length, as we can see also in Fig \ref{Fig:level_pattern}.

The center of mass was allowed to vary in height, and this has demonstrated to be very useful to achieve bigger steps without reaching joint limits as the robot stretches the stance leg while the other one is moving forward.

With 4\si{\second} the walking motion is stable, as the CoM can be quite closely tracked and the ZMP always lies inside the support polygon as shown in Fig \ref{Fig:level_pattern}. The current consumption in this case is higher on the hip roll and knee motors, as they are the most demanding ones during swing phases where the whole weight of the robot is on a single leg.

By further decreasing the time the walking could not be achieved anymore due to high destabilization fo the motion.
We have computed the CoM, ZMP and joint angles tracking errors, and have observed that there is an increase in the errors with the increase of walking velocity.
In the case of stable walkings the average CoM tracking error is in the range of $0.03\sim0.04$\si{\meter}, and average ZMP tracking error of $0.08\sim0.09$\si{\meter}. The joints tracking error shows an increase with the reduction of stride time, ranging from 0.57 \si{\degree} in the 8\si{\second} case up to 2\si{\degree} in the 4\si{\second} case.
In the case of walking with 3\si{\second} the errors have been much higher, mainly in the joint tracking error, which is of 6\si{\degree} in average. This led to very unstable motions that brought the CoM far from the desired one and consequently the ZMP out of the support area, from which the robot could not recover.

\begin{table}
 \begin{center}
 \caption{KPIs in level ground walk}
 \begin{tabular}{|c|c|}
  \hline
   Cost of transport $E_{CT}$ & 4.27 \\
  \hline
   Maximum velocity $v_{max}$ & 0.037 [m/s]\\
   \hline
   Froude number $Fr$ & 0.0165 \\
   \hline
    & CoM$_e$ = 0.038 [m]\\
   Execution precision & ZMP$_e$ = 0.09 [m]\\
   & $q_e$ = 2.11 [deg] \\
   \hline
 \end{tabular}
 \label{Tab:kpi_level}
 \caption*{\small{Measurements are referred to the case as in Fig. \ref{Fig:level}.}}
 \end{center}
\end{table}

\subsection{Slope}
\begin{figure}[!tbp]
    \centering
    \begin{subfigure}[b]{0.5\textwidth}
      \includegraphics[scale = 0.2]{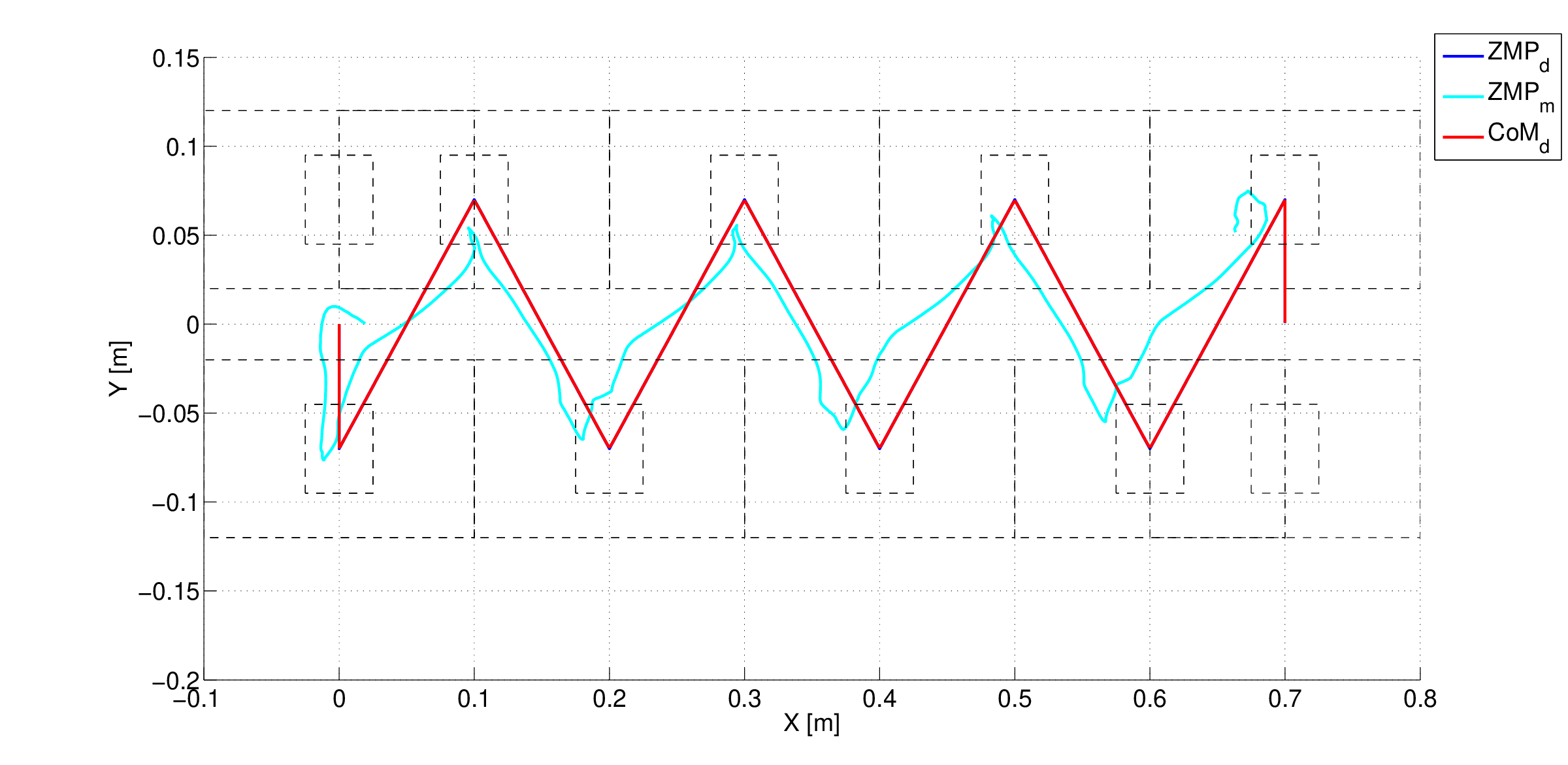}
      \caption{Feet pattern with ZMP and CoM}
      \label{Fig:slope_pattern}
    \end{subfigure}

    \begin{subfigure}[b]{0.5\textwidth}
      \includegraphics[scale = 0.15]{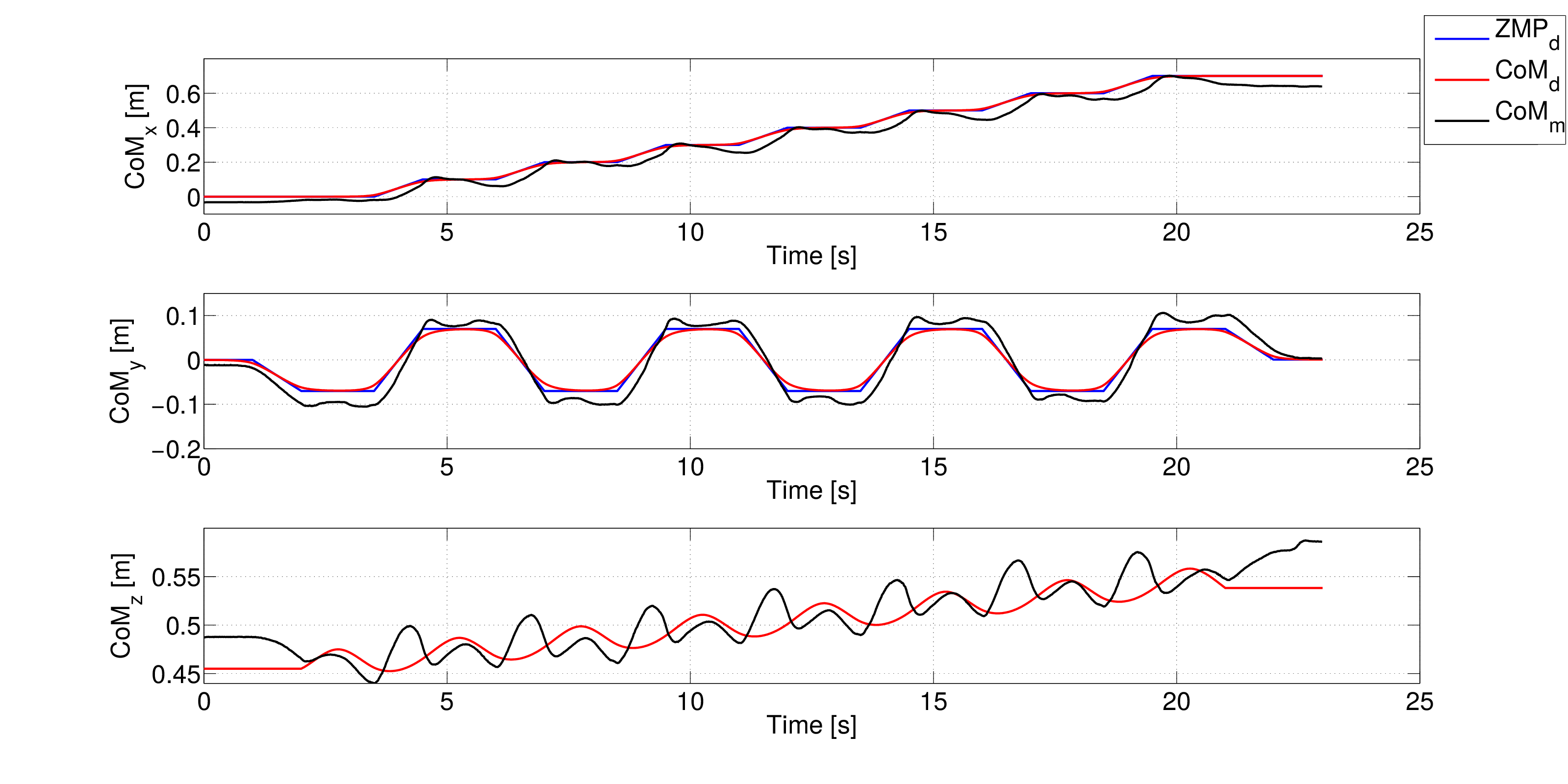}
      \caption{ZMP and CoM trajectory}
      \label{Fig:slope_com}
    \end{subfigure}

    \begin{subfigure}[b]{0.5\textwidth}
      \includegraphics[scale = 0.2]{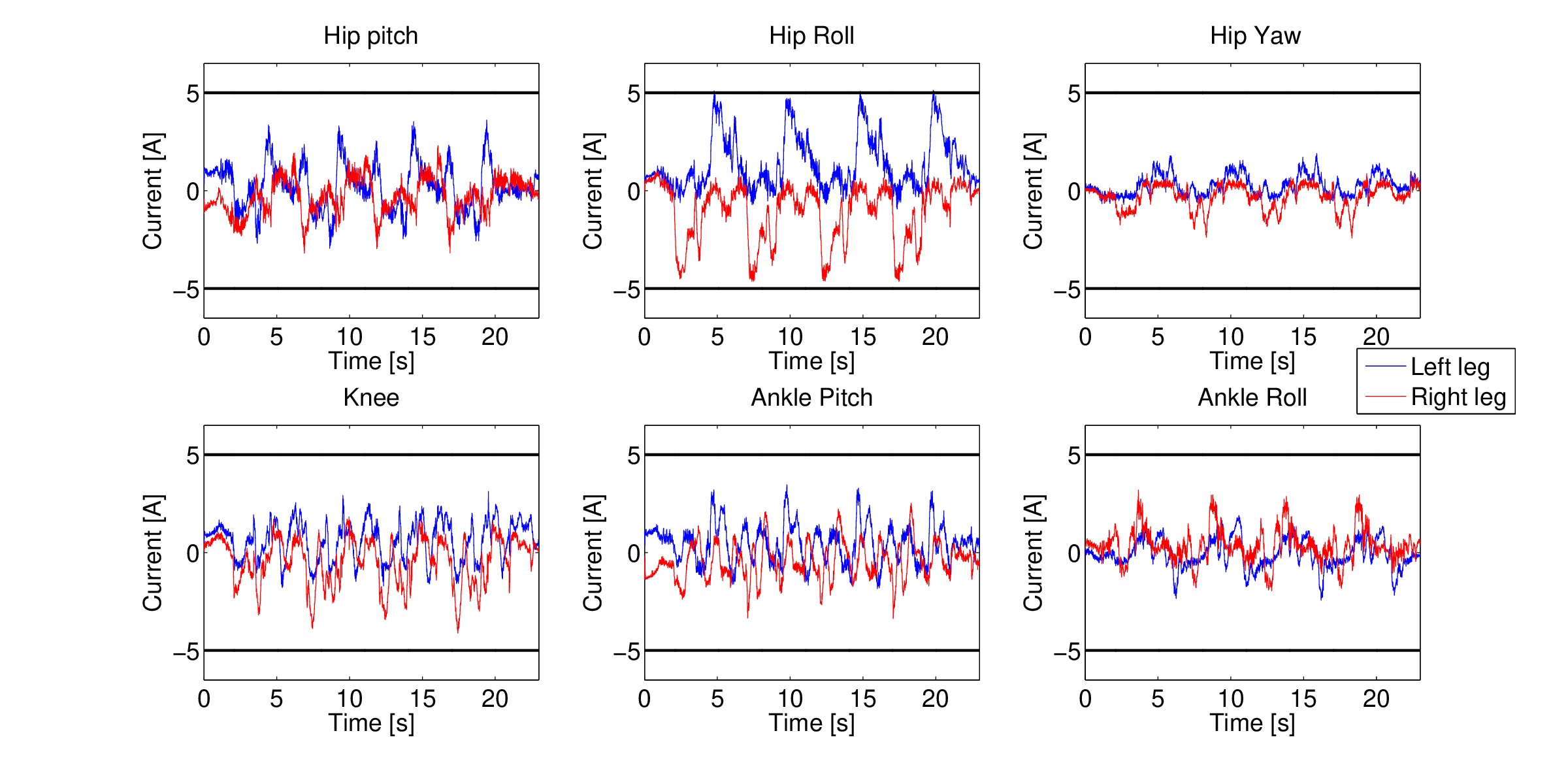}
      \caption{\small{Currents of leg motors, black lines are the default current limits of +/-5 [A].}}
      \label{Fig:slope_current}
    \end{subfigure}
    \caption{\small{Walking up slope with T\_stride 5 [s], step\_length 0.1 [m] and slope inclination of 7 [deg]. Center of mass has a variation of 0.02 [m].}}
    \label{Fig:slope_up}
\end{figure}

\begin{figure}[!tbp]
    \centering
    \begin{subfigure}[b]{0.5\textwidth}
      \includegraphics[scale = 0.2]{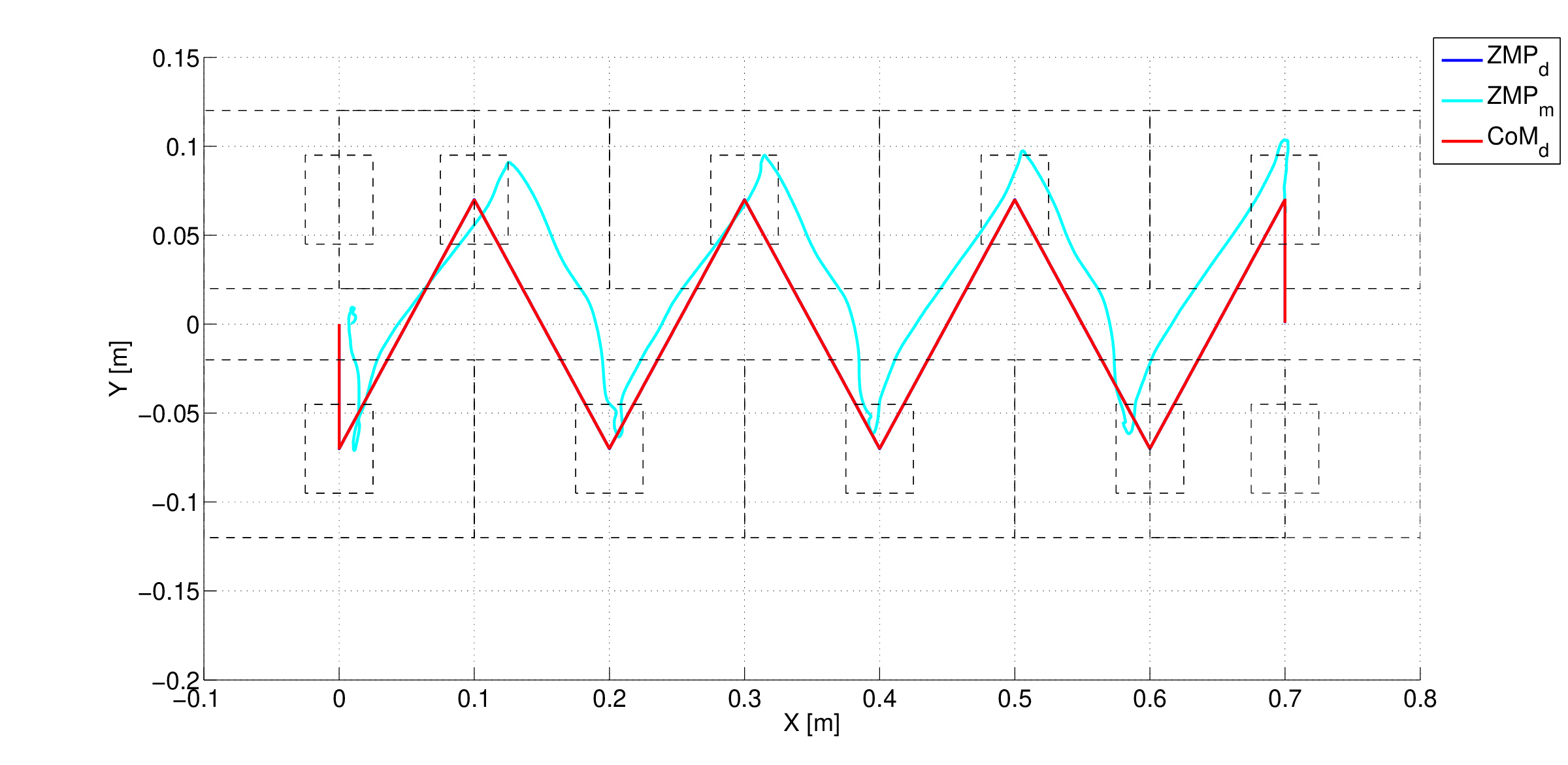}
      \caption{Feet pattern with ZMP and CoM}
      \label{Fig:slope_down_pattern}
    \end{subfigure}

    \begin{subfigure}[b]{0.5\textwidth}
      \includegraphics[scale = 0.15]{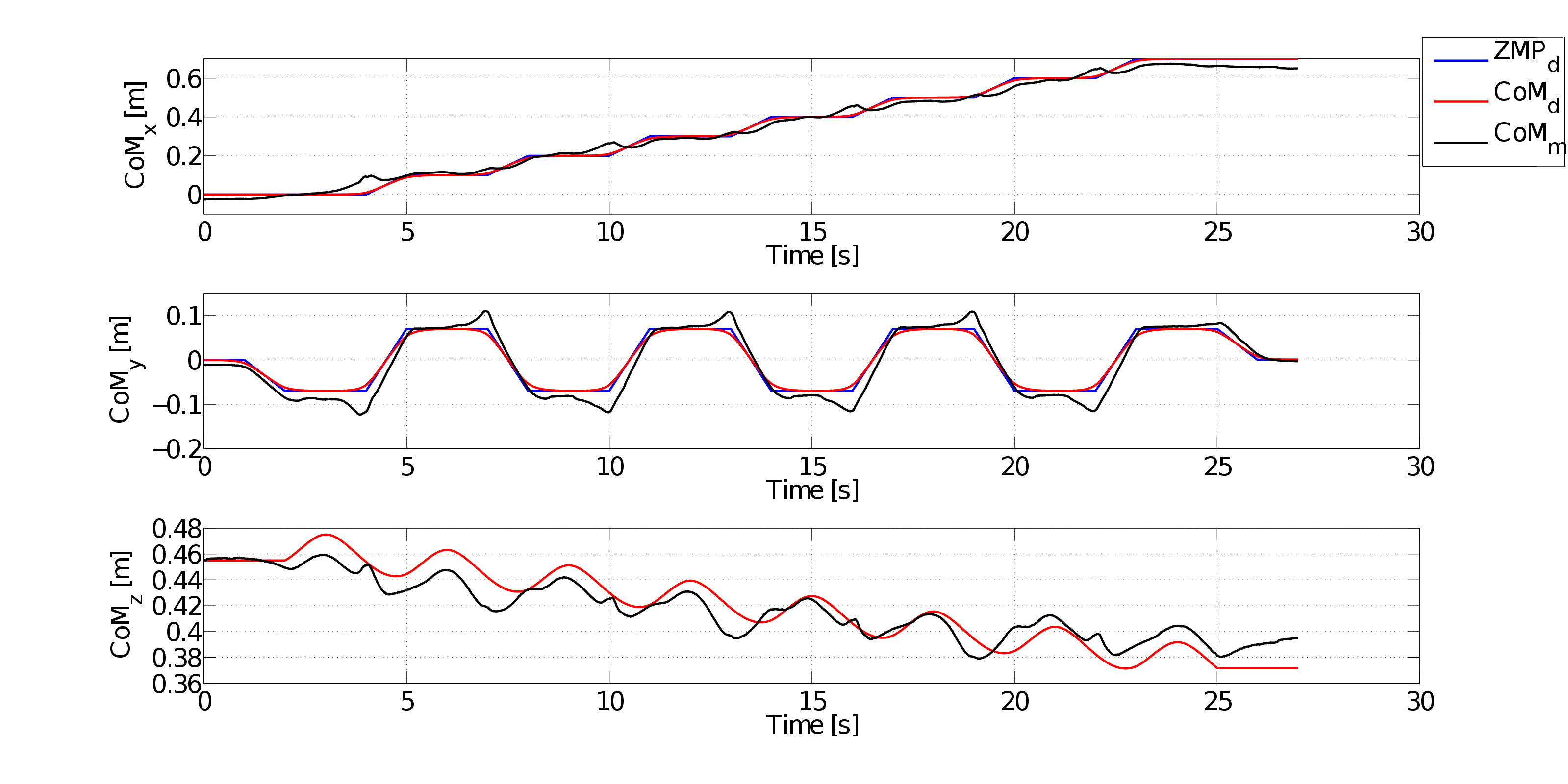}
      \caption{ZMP and CoM trajectory}
      \label{Fig:slope_down_com}
    \end{subfigure}

    \begin{subfigure}[b]{0.5\textwidth}
      \includegraphics[scale = 0.2]{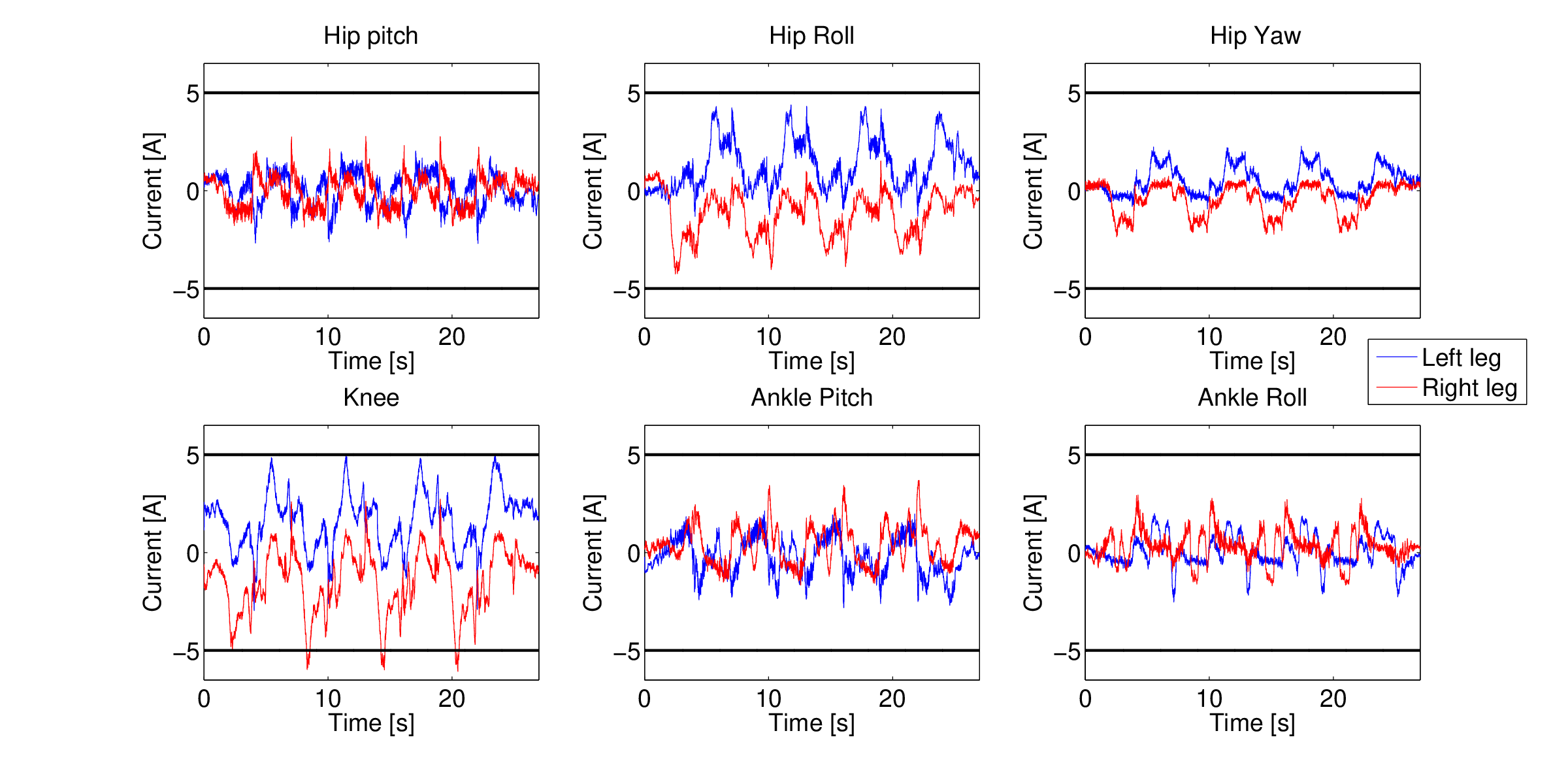}
      \caption{Currents of leg motors, black lines are the default current limits of +/-5 [A].}
      \label{Fig:slope_down_current}
    \end{subfigure}
    \caption{\small{Walking down slope with T\_stride 6 [s], step\_length 0.1 [m] and slope inclination of -7 [deg]. Center of mass has a variation of 0.02 [m].}}
    \label{Fig:slope_down}
\end{figure}

We have performed experiments with both up and down slopes with two different inclinations and a initial constant step length of 0.05\si{\meter}, increased then to 0.1\si{\meter}. The step length has been increased seeing that the motion with smaller step length has been proven to be very stable.

The upslope motion has been demonstrated to be stable with inclination of 7\si{\degree} and a step time of 5\si{\second}, however we can see in Fig \ref{Fig:slope_pattern} that the CoM and the ZMP are not tracked as closely as in the level ground case. In particular we can observe that the ZMP is shifted backwards due to the robot slipping slightly downslope. In fact the success rate is of 40\% only, against the much more stable case of 6\si{\second}.
From the CoM, ZMP and joint angles tracking errors we could observe that the errors are in average higher than the level ground case. In particular the CoM tracking error is in the range of $0.04\sim0.05$\si{\meter} and ZMP tracking error of $0.11\sim0.13$\si{\meter}, while joint angles tracking error of $0.8$\si{\degree}$\sim1$\si{\degree}.

The knee motor requires less current here due to the shorter stride time and the inclination of the slope that allows the knee to have smaller motion range.

In the downslope case we could achieve stable motions with the same inclination of 7\si{\degree}, but with a minimum stride time of 6\si{\second}, where we have achieved only 20\% success rate. As shown in Fig. \ref{Fig:slope_down_pattern}, the ZMP is slightly shifted forward, for the same reason as in the upslope case. 
In this case the tracking errors are smaller than the upslope case, however the motion is also slower.
The CoM tracking error is in the range of $0.4\sim0.5$\si{\meter}, the ZMP tracking error of $0.08\sim0.1$\si{\meter} and the joint tracking error of $0.8$\si{\degree}$\sim1$\si{\degree}.
Stable motions with smaller stride time have not been achieved successfully because downslope situations are more difficult to handle given that the robot shifts the weight forward and the gravity and slippage effects 
destabilize the motion. Knee motor current is also more demanding as the knee joint has to bend more to keep the weight on the stance foot while swinging the other one towards a lower position. Downslope walking is, in fact, also demanding for the knee joint in humans.

It should be noted that walking on slope is in general less stable than the level ground case due to gravity and slippage effects, and therefore tracking errors, mainly of CoM and ZMP, are bigger.
With a slope inclination of 7\si{\degree} the ankle pitch joint is very close to its limits. With proper control of the robot a slightly higher inclination might be achieved, however the mechanical limitation does not allow stable motions in higher inclinations.

\begin{table}
 \begin{center}
 \caption{KPIs in slope walk up}
 \begin{tabular}{|c|c|}
  \hline
   Cost of transport $E_{CT}$ & 4.27 \\
   \hline
   Maximum velocity $v_{max,}$ & 0.03 [m/s]\\
   \hline
   Maximum angle $\alpha_{up}$ & 7 [deg] \\
   \hline
    & CoM$_e$ = 0.053 [m]\\
   Execution precision & ZMP$_e$ = 0.12 [m]\\
   & $q_e$ = 1.03 [deg] \\
   \hline
 \end{tabular}
 \label{Tab:kpi_level}
 \caption*{\small{Measurements are referred to the case as in Fig. \ref{Fig:slope_up}.}}
 \end{center}
\end{table}
\begin{table}
 \begin{center}
 \caption{KPIs in slope walk down}
 \begin{tabular}{|c|c|}
  \hline
   Cost of transport $E_{CT}$ & 5.61 \\
  \hline
   Maximum velocity $v_{max,}$ & 0.026 [m/s]\\
   \hline
   Maximum angle $\alpha_{down}$ & -7 [deg] \\
   \hline
    & CoM$_e$ = 0.048 [m]\\
   Execution precision & ZMP$_e$ = 0.09 [m]\\
   & $q_e$ = 0.88 [deg] \\
   \hline
 \end{tabular}
 \label{Tab:kpi_level}
 \caption*{\small{Measurements are referred to the case as in Fig. \ref{Fig:slope_down}.}}
 \end{center}
\end{table}

\subsection{Stairs}
\begin{figure}[!tbp]
    \centering
    \begin{subfigure}[b]{0.5\textwidth}
      \includegraphics[scale = 0.2]{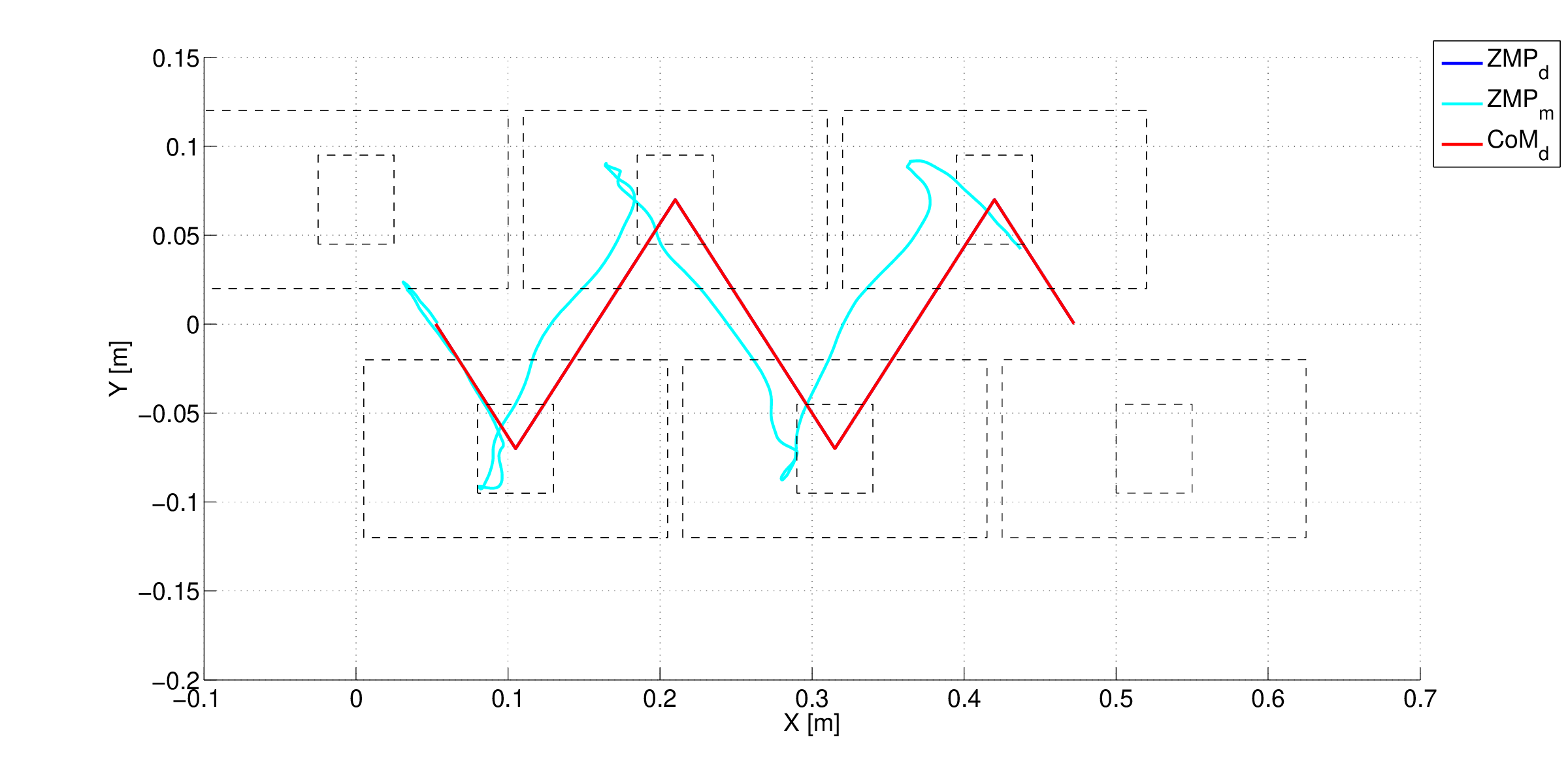}
      \caption{Feet pattern with ZMP and CoM}
      \label{Fig:stair_pattern}
    \end{subfigure}

    \begin{subfigure}[b]{0.5\textwidth}
      \includegraphics[scale = 0.15]{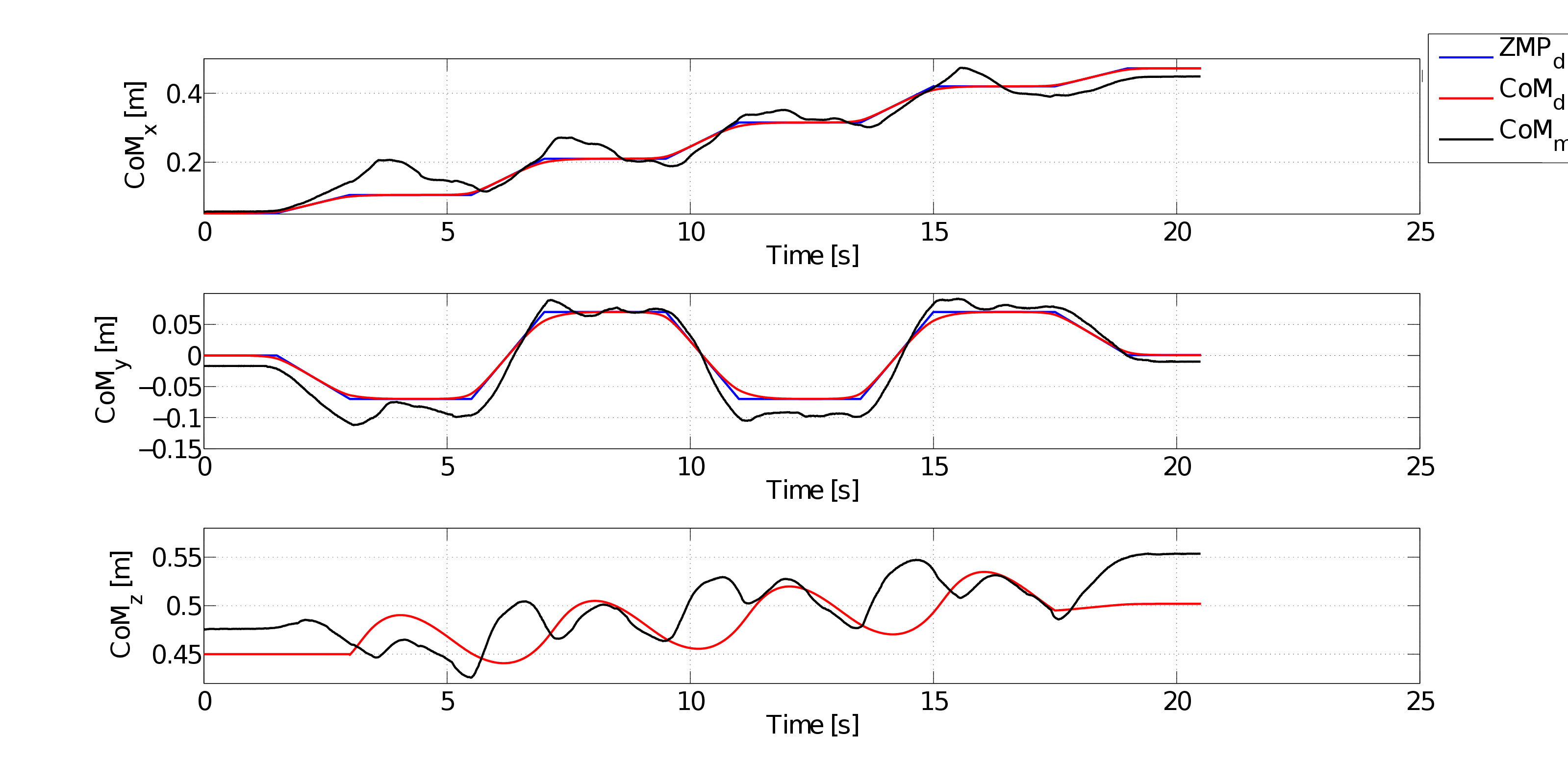}
      \caption{ZMP and CoM trajectory}
      \label{Fig:stair_com}
    \end{subfigure}

    \begin{subfigure}[b]{0.5\textwidth}
      \includegraphics[scale = 0.2]{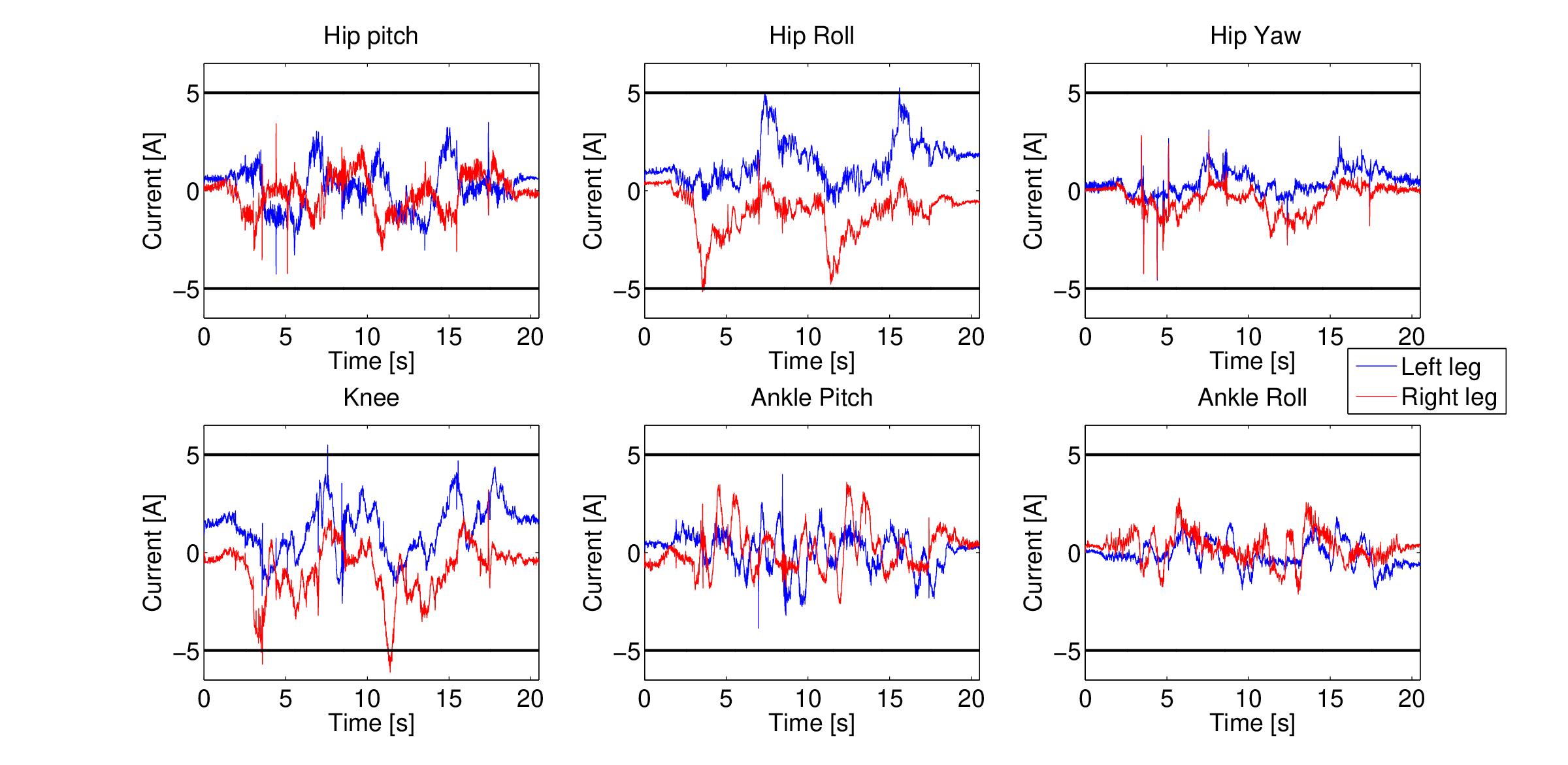}
      \caption{\small{Currents of leg motors, black lines are the default current limits of +/-5 [A].}}
      \label{Fig:stair_current}
    \end{subfigure}
    \caption{\small{Walking up stairs with T\_stride 8 [s], stair\_height 0.02 [m] and stair\_length 0.21 [m]. Center of mass has a variation of 0.025 [m].}}
    \label{Fig:stair}
\end{figure}

It is a known fact, also from human walking, that stair climbing is demanding for the knee joint.
For this reason, in the case of stair walking we tested first for a height of 0.01\si{\meter}, where we had experienced current limits on the knee motor. To successfully perform the motion the limit has been increased up to 5.5\si{\ampere}. We achieved walking on stairs of 0.02\si{\meter} high, by increasing the knee current limit up to 6.5\si{\ampere}, as we can see from Fig. \ref{Fig:stair_current}. We decided to not increase further the stair height and the walking velocity for safety reasons.

Stair walking has been the one with highest CoM and ZMP tracking errors. While the average joint angles tracking error is of 0.9\si{\degree}, the CoM tracking error reached up to 0.09\si{\meter} and the ZMP tracking error up to 0.14\si{\meter}. Proving that in all the tested environments, stair climbing is the most challenging.

As most of the walking humanoid robots, iCub also has legs that are short with respect to its feet size. This means that in stair walking, it is kinematically possible to perform a stair per step as humans do, but it would be dynamically demanding resulting a in a very high current required for the knee motor. In order to avoid this, the robot should perform one stair per time, i.e. put both feet on the stair at every second step.
In our case, in order to achieve a more ``human-like'' stair walk, we created a scenario in which the stairs are shifted, as we can see from Fig. \ref{Fig:scenario}c and \ref{Fig:stair_pattern}. This means that except for the first and last steps, where the feet of the robot are aligned on the same level, the actual height performed is double of the step height (0.02 or 0.04\si{\meter}).

In walking up the stairs the feet ankle pitch is very close to its joint limit, which represents an issue in the case of walking down the stairs. In the latter case, in single support phase, the ankle pitch of the stance foot has to bend to a very small angle (motion has to be large), as the foot has to stay flat on the floor. As it is for the time being unfeasible, the motion has not been performed.

\begin{table}[!tbh]
 \begin{center}
 \caption{KPIs in stair climbing}
 \begin{tabular}{|c|c|}
  \hline
   Cost of transport $E_{CT}$ & 5.06 \\
   \hline
   Maximum step height $h_{up}$ & 0.02 [m]\\
   \hline
    & CoM$_e$ = 0.09 [m]\\
   Execution precision & ZMP$_e$ = 0.14 [m]\\
   & $q_e$ = 0.9 [deg] \\
   \hline
 \end{tabular}
 \label{Tab:kpi_level}
 \caption*{\small{Measurements are referred to the case as in Fig. \ref{Fig:stair}.}}
 \end{center}
\end{table}

\subsection{Summary}
All the performed motions are stable within the limitations of the robot and the applied methodology.
We could gather interesting information about the walking capabilities of iCub and measured the KPIs that can serve as reference for future improvements.

From all the analyzed walking scenarios, it is possible to summarize that the joint limits of the iCub are conservative for walking in environments different from flat ground.
Mostly the limit lies in the ankle pitch joint motion range. Given that the feet of the robot are one single block with rigid flat surface, without the possibility of tip toe flexibility, there is a necessity of increasing this limit.

A possible way to cope with the current situation without mechanically modifying the robot is to build soles of different shapes for the robot. For instance, in the case of walking down stairs or up slope, the issue lies in the limited angle of the foot while bending towards the lower leg, while the joint does not need to bend in the other direction. In this case it is possible to build a slope shaped sole (a wedge) for the robot, such that the zero position of the ankle pitch is shifted, increasing the motion range. This might introduce further problems, as the height of the robot increases and the range of the ankle pitch is reduced in the opposite direction.
Furthermore this method cannot be used in walking down a slope, where the opposite problem exists, and further investigations are necessary to verify if with such external sole the robot is still capable of level ground walking. Nevertheless it is worth testing in further experiments.

\section{CONCLUSIONS}
We illustrated and analyzed the performances of the iCub robot with the reduced version HeiCub by using an offline ZMP based pattern generator and an IPOPT based inverse kinematics module.

The software module as illustrated in section \ref{sec:workflow} can be used to test also other walking environments that do not expect external perturbations, such as beam, step stones, curved walk etc.
It allows a fast check of the wanted/expected motion on the simulator or the real robot, to gain useful information to be used in future implementations of control methods such as the ones we discussed in the paper.
It also can serve as a trajectory generator, where the trajectory can be then fed to a torque control strategy.

The success of stable walking with the adopted method relies highly on the precision of the execution of the motion with position control, the fine calibration of the relative encoders and the precision of the kinematic model used in the IK. This means that the performances depend also on the fine tuning of the position PID (Proportional, Integral, Derivative) gains.

We have however achieved our goals of implementing walking on the iCub and exploiting its capabilities and limitations in the illustrated scenarios and proposed a possible solution for the restricted joint limits. However an issue that cannot be solved is the current of the motors. The motors can sustain only up to a certain current level, which means that it is not possible to perform challenging motions where high power is required, even though the motion would be kinematically possible. This of course is an exposed limitation when doing not only position-controlled tasks, but more advanced torque-controlled ones, that can now be anticipated through this study.

In future work we will perform walking with online pattern generator using as feedback data gathered through sensors of the robot, with a more accurate floating base estimation module, e.g. using motion capture systems. We will also perform walking with the SEA modules mounted in the knees and ankles, differently than the present work, by means of model-based optimal control taking into account the elasticity properties of the joints. The performance indices presented in this paper will thus serve as a basis of comparison for possible improvements.






\bibliographystyle{orbref-num}
\bibliography{CompliancePapers}

\end{document}